# Deep Learning Approach for Enhanced Transferability and Learning Capacity in Tool Wear Estimation

Zongshuo Li[a,*], Markus Meurer[a], Thomas Bergs[a,b]

[a]*Laboratory for Machine Tools and Production Engineering (WZL) of RWTH Aachen University, Campus-Boulevard 30, 52074 Aachen, Germany*
[b]*Fraunhofer Institute for Production Technology IPT, Steinbachstr. 17, 52074 Aachen, Germany*

* Corresponding author. Tel.: +49-241-80-20522; *E-mail address:* Z.Li@wzl.rwth-aachen.de

**Abstract**

As an integral part of contemporary manufacturing, monitoring systems obtain valuable information during machining to oversee the condition of both the process and the machine. Recently, diverse algorithms have been employed to detect tool wear using single or multiple sources of measurements. In this study, a deep learning approach is proposed for estimating tool wear, considering cutting parameters. The model's accuracy and transferability in tool wear estimation were assessed with milling experiments conducted under varying cutting parameters. The results indicate that the proposed method outperforms conventional methods in terms of both transferability and rapid learning capabilities.

## 1. Introduction

The assessment of tool wear is crucial in manufacturing systems as it enables timely decisions for tool replacement, ensures workpiece quality, and enhances productivity while yielding cost savings. In recent years, advances in computer, sensor, and signal processing technologies have facilitated real-time monitoring of machining process signals, promoting the development of data-driven approaches for tool wear estimation [1,2]. The primary concept behind data-driven methods involves training a tool wear model using historical data, which is then utilized to estimate tool wear by inputting online-collected data into the trained model.

### 1.1. Tool wear estimation with machine learning

In initial research, a variety of traditional machine learning algorithms were employed for estimating tool wear, such as random forest (RF) [3,4], support vector machine (SVM) [5,6], gaussian process regression (GPR) [7,8], artificial neural network (ANN) [9,10], fuzzy neural network (FNN) [11], and others. These methods generally necessitate manual feature extraction and selection, which becomes unfeasible with the growing amount of monitoring data, resulting in decreased estimation accuracy. The significant progress in computational power has enabled the use of deep learning algorithms for tool wear estimation due to their inherent ability to learn feature representations [12,13]. The representing deep learning algorithm, convolutional neural network (CNN) [14] is widely employed for processing not only image data but also sequential data [15,16,17]. CNN exhibits weight sharing, local connectivity translational invariance, and spatial invariance properties, by stacking convolutional and pooling layers. As a result, it is frequently employed to extract one-dimensional or two-dimensional spatial features from force or vibration signal sequences. Martínez-Arellano et al. [18] present a new

**Nomenclature**

| | |
|---|---|
| Conv. | Convolution |
| Cond. | Condition |
| $v_c$ | Cutting speed |
| $a_p$ | Cutting depth |
| $a_e$ | Cutting width |
| FPT, $f_z$ | Feed per tooth |
| $M_{Spindle}$ | Spindle torque |
| $F_{RCD\_x}, F_{RCD\_y}$ | Orthogonal cutting force components from the rotating cutting dynamometer |
| $F_{RCD\_resultant}$ | Resultant cutting force from the rotating cutting dynamometer |
| $F_{SD\_x}, F_{SD\_y}$ | Orthogonal cutting force components from the stationary dynamometer |
| $F_{SD\_resultant}$ | Resultant cutting force from the stationary dynamometer |
| $F_{SD\_feed}$ | Cutting force from the stationary dynamometer aligned with the feed direction |
| $F_{SD\_normal}$ | Cutting force from the stationary dynamometer perpendicular to the feed direction |
| $VB_1$ | Average width of flank wear land in section one |
| $VB_{max\_1}$ | Maximum width of flank wear land in section one |
| $VB_{E1}$ | Average width of flank wear land of cutting edge one |
| $VB_{max\_E1}$ | Maximum width of flank wear land of cutting edge one |

big data approach for tool wear classification using CNN with signal imaging, and achieve a classification accuracy of tool condition above 90%. Huang et al. [19] introduce a multi-scale convolutional neural network featuring an attention fusion module for tool wear classification, demonstrating the effectiveness and superiority of the proposed method. These models demonstrate powerful spatial and temporal feature extraction abilities.

*1.2. Transferring the tool wear estimation model*

During practical applications, cutting conditions change due to alterations in the workpiece, cutting tool, or cutting parameters. Such changes result in significant variations in monitoring data distributions. The network obtained above is designed for a single cutting condition, and collecting sufficient data samples for each condition to train a deep learning model entails considerable time, experimentation, and labor costs. Consequently, the aforementioned method faces challenges in precisely estimating tool wear under varying cutting conditions. Li et al. [20] utilize the entropy weight method and gray correlation analysis to identify signal features exhibiting strong relationships with tool wear and weak relationships with cutting conditions. This strategy minimizes the influence of cutting conditions on both monitoring signals and tool wear, enabling the model to adapt to other parameter variations and preserve estimation accuracy. Nonetheless, the feature engineering involved in this approach becomes unfeasible when confronted with high-frequency, large-volume industrial data. The re-selection of features and manual examination under varying cutting conditions are still inevitable, and there is potential information loss in comparison to deep learning methods based on raw data.

This study aims to introduce a deep learning approach that considers cutting conditions, which could substantially enhance the estimation of actual tool wear under diverse cutting conditions compared to previous methods. To assess the model's accuracy and transferability, milling experiments were conducted under various cutting parameters. The subsequent chapter outlines the model's structure. Chapter 3 details the experimental setup, data processing flow, and validation strategy employed to evaluate the model. Chapter 4 offers an analysis and discussion of the validation results. The final chapter concludes by summarizing the proposed approach's characteristics and providing a perspective on its future advancements.

**2. Incorporating Cutting Conditions in Model**

The objective of this study is to ascertain the enhanced transferability of the proposed deep learning method in contrast to approaches that disregard the cutting condition, thereby emphasizing the model's comparative performance superiority. The scope of this paper does not encompass further optimization strategies, such as the employment of more sophisticated residual networks or the integration of recurrent neural networks, including Long Short-Term Memory (LSTM) architectures. The overarching concept of the proposed deep learning method is depicted in Fig. 1.

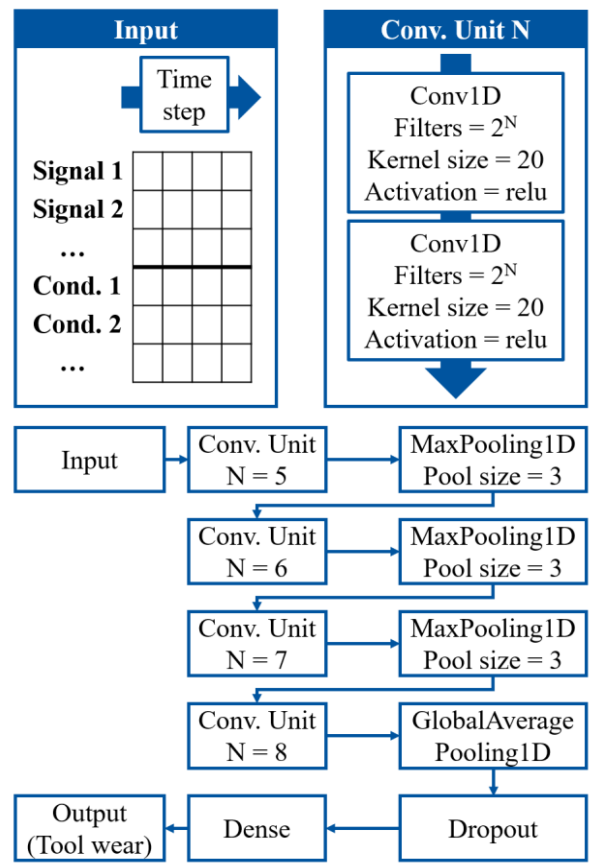

Fig. 1. Concept of the proposed deep learning method

The model consists of a CNN designed to estimate tool wear under specific cutting conditions. A typical CNN architecture comprises multiple distinct kernel filters that effectively and autonomously extract highly discriminative features from the input data, encompassing both the time domain of individual time series signals and the cross-series domain. The fundamental structure of CNN primarily includes an input layer, convolutional layers, pooling layers, fully connected layers, and an output layer. In the input layer, each time series signal from the sensor serves as a signal channel, with each cutting condition incorporated as an additional separate channel. Consequently, the data at each time point comprises the current value of each time series signal as well as the prevailing cutting condition such as feed per tooth, cutting depth, etc. The initial convolutional layer emphasizes correlation and fusion among the disparate time series signals, accompanied by minimal compression in the time domain. Subsequent convolutional layers concentrate on the time domain. The output layer is configured with multiple outputs, aiming for multi-scale tool wear measurements to enhance model performance and robustness. The specific input signals and tool wear measurements are delineated in Chapter 3. The training configurations for all models are consistent across both the test and reference models. The model employs a learning rate decay strategy, incorporating an initial learning rate of 0.001, a decay rate of 0.7, and a step decay approach with decay applied every 20 epochs. Utilizing the Adam optimization algorithm, each model undergoes training for a total of 100 epochs.

## 3. Experimental Setup for Model Validation

The subsequent sections elucidate the experimental setup, encompassing the tools, materials, and sensors employed, in addition to the data processing and model validation methodologies implemented.

### 3.1. Experimental setup

In order to train the proposed method and validate its transferability, 13 groups of milling experiments were designed with different process parameters and conducted utilizing a 5-axis machining center DMU 85 Monoblock from DMG Mori. Fig. 2 shows the experimental setup. An Inconel 718 DA ring-shaped workpiece was fixed on the stationary dynamometer, with a height of 3 cm, an outer diameter of 30 cm, and an inner diameter of 8 cm. Uncoated end mills were employed, featuring four cutting edges, a diameter of 6 mm, and a corner radius of 0.15 mm. The cutting edges remained unrounded and tool substrate is HM-MG10. Throughout the experiments, the milling process was executed circumferentially on the workpiece, with pre-determined cutting parameters. The duration of each cut, ranging from 10 to 20 seconds, was contingent upon specific cutting parameters and the progression of tool wear during the examination. Following each cut, the tool wear was assessed with a Keyence VHX 900F microscope. When the maximum width of flank wear land $VB_{max}$ reached 200 μm on any of the four cutting edges, the tool was deemed entirely worn and subsequently not used further. Parameters monitored during the cutting process included cutting force components, spindle torque, and selected internal machine signals. A Kistler 9255C stationary dynamometer (SD) was employed to measure the workpiece-side cutting force components. Meanwhile, a Kistler 9170A rotating cutting dynamometer (RCD) was utilized to measure spindle torque and tool-side cutting force components. Additionally, auxiliary drive positions were documented and synchronized with the stationary dynamometer and RCD readings.

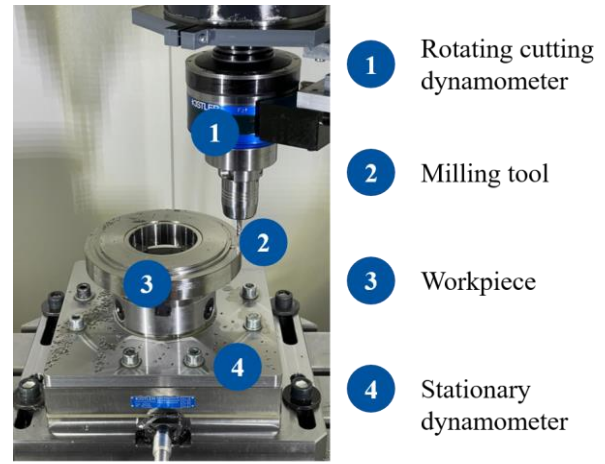

Fig. 2. Experimental setup

Regarding the process parameters, all experiments will be conducted at a consistent cutting speed $v_c$ of 25 m/min, a cutting depth $a_p$ of 2.5 mm, and a cutting width $a_e$ of 1.5 mm. The variability in parameters stems from the feed per tooth $f_z$ (FPT), which ranges between 0.015 mm and 0.060 mm. Table 1 presents a summary of the investigated feed per tooth values and their corresponding tool numbers.

Table 1. Feed per tooth values for milling experiments

| Tool Nr. | Feed per tooth $f_z$ [mm] | Cutting time per cut [s] |
|---|---|---|
| 1, 2, 3 | 0.015 | 15-25 |
| 4, 5, 6 | 0.030 | 12-22 |
| 7, 8, 9 | 0.045 | 14-17 |
| 10, 11 | 0.525 | 13 |
| 12, 13 | 0.060 | 12 |

## 3.2. Data preparation

To model tool wear using the data-driven approach, the data collected during the tests must first be processed. This includes data preprocessing and labeling. Fig. 3 shows the procedure for data pre-processing. A single cutting operation comprises three distinct phases: cut entry, milling, and cut exit, with the milling process characterized by consistent cutting conditions. The initial step involves isolating the signal segments corresponding to the milling process based on the auxiliary drive position. Each signal segment encompasses five signal channels: the spindle torque ($M_{Spindle}$), two orthogonal cutting force components from the RCD ($F_{RCD\_x}$ and $F_{RCD\_y}$) and two orthogonal cutting force components ($F_{SD\_x}$ and $F_{SD\_y}$) from the stationary dynamometer. The orthogonal cutting force components $F_{SD\_x}$ and $F_{SD\_y}$ are transformed, utilizing the auxiliary drive position, into a cutting force aligned with the feed direction ($F_{SD\_feed}$) and a cutting force perpendicular to the feed direction ($F_{SD\_normal}$). This conversion facilitates the representation of circular motion as linear motion, enabling the comparison of signals at varying positions. The resultant forces, $F_{RCD\_resultant}$ and $F_{SD\_resultant}$, can be calculated from $F_{RCD\_x}$ and $F_{RCD\_y}$ or $F_{SD\_feed}$ and $F_{SD\_normal}$, respectively. Consequently, seven sensor signal series can be obtained from a single cutting operation, providing comprehensive data for analysis and optimization of the milling process. Subsequently, the final one-second interval of the seven-channel raw sensor data is extracted as the representative signal for each cut, due to its proximity to the tool wear measurement. Upon conducting a frequency analysis of all signals, the highest frequency is found to be approximately 4000 Hz. Consequently, a low-pass filter with a cutoff frequency of 8000 Hz is employed to reduce noise within the signals. Given that the sampling time is 50 µs, this one-second duration comprises 20,000 time steps. Correspondingly, with seven sensor signals, this amounts to a total of 140,000 data points.

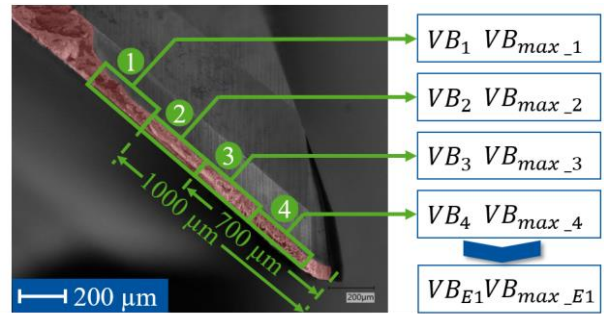

Fig. 4. Measurement results for one of the cutting edges

The width of flank wear land is used as the output of the model. To generate the label, the cutting edge was divided into four sections along its length. The first section extended from 1000 to 1300 µm from the edge's end, the second from 700 to 1000 µm, the third from 400 to 700 µm, and the last from 100 to 400 µm. For each section, the average and maximum width of flank wear land (e.g. $VB_1$ and $VB_{max\_1}$) were determined. In addition, for each cutting edge, the overall average and maximum width of flank wear land (e.g. $VB_{E1}$ and $VB_{max\_E1}$) were calculated, resulting in 10 values per cutting edge. Fig. 4 presents an illustration of 10 measurement results for one of the cutting edges. By averaging the 40 widths of flank wear land from the four cutting edges of a single tool by cutting edge, a final set of ten widths of flank wear land was obtained. The rationale behind this procedure is to augment the learning reference target of the model, thereby enhancing its overall performance.

## 3.3. Model validation methodologies

The validation scenarios were categorized into two types: transfer to a single new FPT and transfer to two new FPT. Each of these scenarios considered whether the data of a tool under the same FPT is included in the training dataset, in order to verify the model's rapid learning capability. Fig. 5 shows examples of each validation scenario. In accordance with various validation scenarios, data of 13 tools encompassing five distinct FPT were divided into a training dataset for model training and a test dataset for validating the model's transfer performance on other FPT. Evaluation metrics, such as root-mean-square error (RMSE) and coefficient of determination ($R^2$), were calculated to assess the model's performance based on the tool wear estimates and actual measurements of the test dataset. A traditional data-driven model was used as a reference, with consistent architecture and training hyper-parameters, but without cutting conditions in the input.

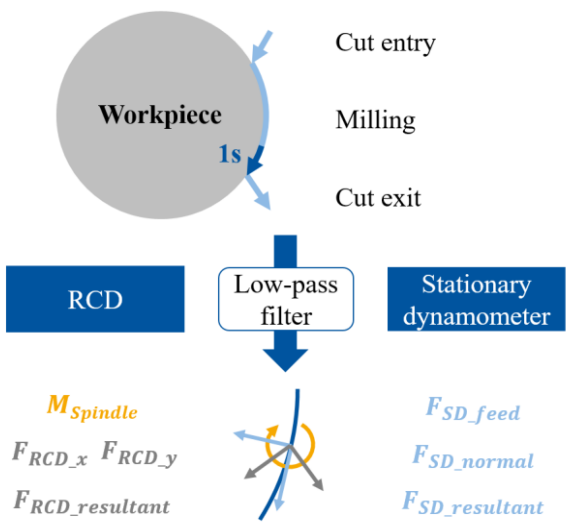

Fig. 3. Procedure for data pre-processing

Fig. 5. Examples of four scenarios to validate transferability

## 4. Results and discussion

In the experiment, a total of 172 milling operations with tool wear measurements were conducted on 13 distinct tools. This chapter assesses the transferability and rapid learning capabilities of the model by comparing the tool wear estimation performance of the proposed model, which takes cutting parameters into account (Test model), against the traditional model that does not consider cutting parameters (Reference model). The average width of flank wear land, one of the model's outputs, is determined by aggregating data from 16 sections of the four cutting edges on each tool. Due to its comprehensive nature, this value is employed for model evaluation.

### 4.1. Transfer to a single new FPT

Fig. 6 illustrates the tool wear estimation performance of the two models when utilizing data from different FPTs as the test dataset. Owing to the incorporation of FPT as input to the test model, it achieves a performance improvement in terms of the metrics (i.e., RMSE and $R^2$) compared to the reference model. When handling data of a new FPT, the test model demonstrates an average performance advantage of 22.9% on RMSE and 36.6% on $R^2$ relative to the reference model. This advantage escalates to 29.2% on RMSE upon learning a part of the new FPT data. Specifically, with a test dataset with a FPT of 0.015 mm, the test model's performance improves by 60.2% following partial learning, while the reference model only exhibits a 41.3% enhancement. This indicates that the test model possesses better transferability compared to the reference model, regardless of whether data with the same FPT is included as part of the training data. Furthermore, the test model also demonstrates a strong learning capability, rapidly adapting model parameters after learning a part of data with a new FPT, resulting in enhanced tool wear estimation accuracy.

### 4.2. Transfer to two new FPT

Fig. 7 displays the model's performance on data of two FPTs after being trained on data of the other three FPTs. Similar outcomes are observed, with the test model exhibiting better performance relative to the reference model on data of new FPT, possessing a 28.4% advantage before partial learning and a 24.4% advantage after partial learning in terms of RMSE. It is noteworthy that the model's performance is inferior on data with an FPT of 0.0150 mm (or 0.0600 mm) compared to data with an FPT of 0.0300 mm (or 0.0525 mm). This suggests that the greater the discrepancy in FPT between the training and test datasets, the poorer the model performance. However, for data with large differences in FPT, the test model maintains a 24.6% advantage on RMSE before partial learning and an 18.4% advantage after partial learning compared to the reference model. This implies that the test model can still provide more accurate tool wear estimates on data with considerable variation in FPT.

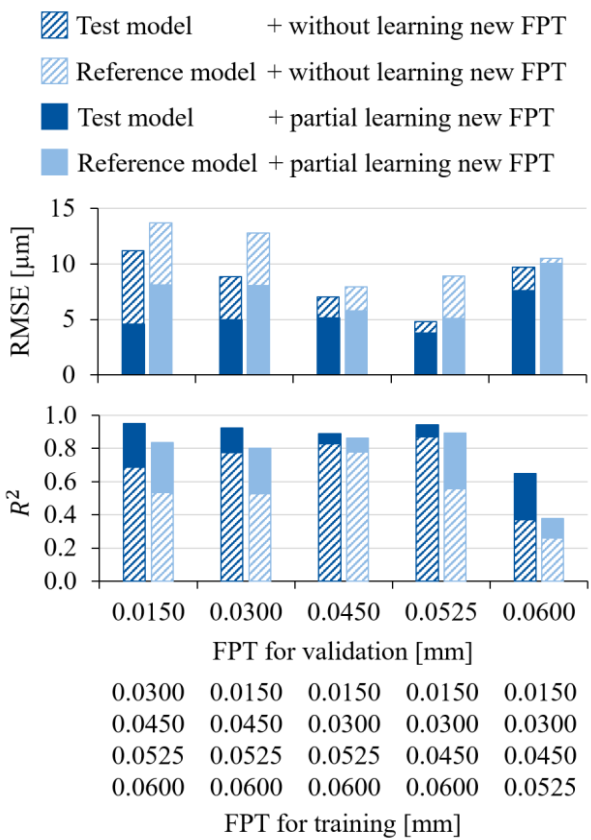

Fig. 6. Model transfer performance to a single new FPT

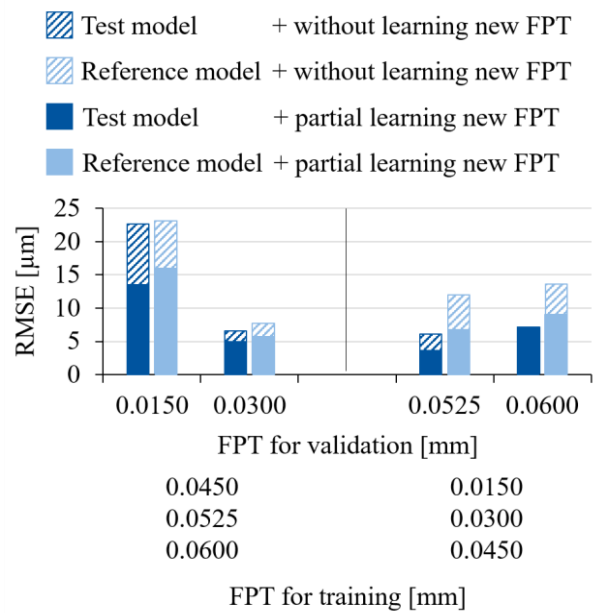

Fig. 7. Model transfer performance to two new FPT

## 5. Conclusion and outlook

Variations in the distribution of monitoring data, stemming from alterations in cutting conditions during machining, pose considerable challenges for conventional models in estimating tool wear under varying cutting conditions. In this study, a deep learning method accounting for cutting conditions is proposed. The model's estimation accuracy and transferability are assessed through milling experiments conducted under diverse FPTs. By incorporating cutting conditions into the model inputs, the proposed approach demonstrates enhanced transferability compared to traditional methods, irrespective of the presence of data with identical FPTs in the training dataset. Moreover, the discrepancy in FPT between the training and test datasets influences the model's transferability. Greater differences yield poorer performance on the data with new FPT. Nonetheless, the proposed method also exhibits a rapid learning capability in comparison to conventional methods, swiftly adjusting model parameters after learning a subset of data with new FPTs, thereby enhancing tool wear estimation accuracy.

In future work, several aspects will be explored to enhance the model's performance. Firstly, the investigated parameter space will be expanded to further validate the model's transferability. Additionally, the model's structure will be optimized, such as integrating CNN and RNN architectures to further improve accuracy. Moreover, an adaptive experimental design algorithm will be proposed to minimize experimental consumption without compromising the model's performance.

**Acknowledgements**

Funded by the Deutsche Forschungsgemeinschaft (DFG, German Research Foundation) under Germany's Excellence Strategy – EXC-2023 Internet of Production – 390621612.